\documentclass[12pt,journal,draftcls,letterpaper,onecolumn]{IEEEtran}
\usepackage{epsfig}
\usepackage{amsfonts}
\usepackage{amssymb}
\usepackage{amsmath}   
\usepackage{latexsym}
\usepackage{graphicx} %
\usepackage{subfigure} 

\usepackage{algorithm}
\usepackage[noend]{algorithmic}
\usepackage{mathrsfs}
\usepackage{type1cm}
\usepackage{eso-pic}
\usepackage{color}
\usepackage{mathabx}
\usepackage{float}
\usepackage{perpage}
\usepackage{rotating}
\usepackage[T1]{fontenc}
\usepackage{authblk}

\begin{document}
\title{A Bayesian Nonparametric Approach to Image  Super-resolution}
\author[1]{Gungor Polatkan}
\author[2]{Mingyuan Zhou}
\author[2]{Lawrence Carin}
\author[3]{David Blei}
\author[4]{Ingrid Daubechies}
\affil[1]{Department of Electrical Engineering}
\affil[3]{Department of Computer Science}
\affil[ ]{Princeton University}
\affil[ ]{Princeton, NJ}
\affil[ ]{\textit {polatkan@princeton.edu, blei@cs.princeton.edu}}
\affil[2]{Department of Electrical Engineering}
\affil[4]{Department of Mathematics}
\affil[ ]{Duke University}
\affil[ ]{Durham, NC}
\affil[ ]{\textit {mz31@duke.edu, lcarin@duke.edu, ingrid@math.duke.edu}}

%
%
%
%
%
%
%
%
%
%
%
%
%
%
%
%
%
%
%
%
%

%
%
%
%

%
%

%
\maketitle

\begin{abstract}
Super-resolution methods form high-resolution images from low-resolution images. In this paper, we develop a new Bayesian nonparametric model for super-resolution. Our method uses a beta-Bernoulli process to learn a set of recurring visual patterns, called dictionary elements, from the data. Because it is nonparametric, the number of elements found is also determined from the data.  We test the results on both benchmark and natural images, comparing with several other models from the research literature. We perform large-scale human evaluation experiments to assess the visual quality of the results. In a first implementation, we use Gibbs
sampling to approximate the posterior.  However, this algorithm is not feasible for large-scale data.  To circumvent this, we then develop an online variational Bayes (VB) algorithm. This algorithm finds high quality dictionaries  in a fraction of the time needed by the Gibbs sampler. \end{abstract}


\section{Introduction}

The sparse representation of signals  with a basis is important in many applications. It has been extensively used in image denoising, inpainting, super-resolution, classification and compressive sensing \cite{KSVD,EladAharon,bachsapirozisserman,bachsapirozisserman2,eladsapiro,eladsapiro2,RanzatoLecun,wrightganeshma}.  

Many real data sets can be sparsely represented in some basis; typically this basis itself has to be learned from the data \cite{KSVD,EladAharon,bachsapirozisserman,eladsapiro2,RanzatoLecun, wrightganeshma,donohosparse,candestao,bachsapiro,RainaBattleLeeNg}. For example,  an image can be represented by weighted combinations of recurrent patterns of pixels. This construction may be beneficial, both while building a model for more accurate representation of the data (e.g. superior image denoising models) and while deriving and implementing an inference procedure for more efficient algorithms.

In this paper we consider image super-resolution (SR), the problem of recovering a high-resolution (HR) image from a low-resolution (LR) image.  It has many applications, e.g., to smart phones, surveillance cameras, medical imaging, and satellite imaging.

There are a variety of approaches for image super-resolution. 
In general, rendering an HR image from an LR
image has many possible solutions. We must use regularization of some
form, i.e., prior information about the HR, to guarantee uniqueness
and stability of the extension. For this purpose, researchers have proposed several
methods \cite{farsiu2004fast, tipping2003bayesian}.
\textit{Interpolation-based methods}, such as the Bicubic method and
Bilinear method, often over-smooth images, losing detail.
\textit{Example-based approaches} use machine learning to avoid this
\cite{freemanSR,kimSR,sun2003image}; they train on ground-truth HR
and LR images, learning a statistical relationship between the two. These relationships are later used to reconstruct unknown HR images from corresponding LR images. Freeman et al. (\cite{freemanSR}) proposes a method that stores a training set of preprocessed patches and uses a nearest-neighbor search to super-resolve. Kim et al. (\cite{kimSR}) proposes using kernel ridge regression with a regularized gradient descent. Another class of SR algorithms use texture similarity  to match image regions with known textures \cite{cohenSR, jiansunSR}. Finally, there are methods for single-image super-resolution.  One classic example is \cite{glasnerSR} which uses recurring patterns at same and different scales in a single image.

In this work, our focus is on SR via example-based  sparse coding. ScSR (Super-resolution via Sparse Representation) is such an algorithm pioneered in \cite{superresolution1}. This algorithm is based on sparse coding via L1 regularized optimization. In \cite{superresolution1},  image data are represented using a collection of dictionary elements (recurring patterns of pixels) that are weighted across different positions. Although very powerful, this model requires one to specify the number of dictionary elements and the variance of the noise model in advance---parameters that may be difficult to assess for real-world images.   It also only provides a batch learning algorithm, i.e., computing model parameters  via a gradient descent algorithm on a fixed small subset of the data. %

 Bayesian nonparametric methods  circumvent all these limitations.  These methods adapt the structure of the latent space to the data and provide a  powerful representation because they infer parameters that otherwise have to be assigned \emph{a priori} \cite{ICML2011Chen_251,VBforIBP,GhahramaniICA,IBP,paisley2009nonparametric,hbp_ibp,zhou2009non, zhou2011dependent,sDDCRP,Ghosh_Sudderth_2012,Kivinen_Sudderth_Jordan_2007}.     The full posterior distribution can be approximated via MCMC or variational inference, yielding sparse representations and learned dictionaries.
  
  Bayesian nonparametric methods have been used in many image analysis applications: to learn deep architectures used for object recognition in \cite{ICML2011Chen_251}, for image inpainting and denoising in \cite{zhou2009non,zhou2011dependent}, for image segmentation   in \cite{sDDCRP,Ghosh_Sudderth_2012}, and to learn  nonparametric multiscale representations of images in \cite{Kivinen_Sudderth_Jordan_2007}.  %

In this paper,  we develop a  Bayesian nonparametric method for super-resolution. We show that inference in  our model  is feasible, performing super-resolution with both a sampling based algorithm and an online variational inference algorithm. In the latter, we approximate the posterior distributions via a stochastic gradient descent over a variational objective that enables us to use the full data set and process the data segment by segment.  We also provide human evaluation experiments which shows that signal-to-noise ratio (a typical quantitative measure of success in image analysis applications) is not necessarily consistent with human judgement.   We devise a new model, new algorithms, and study a human-based evaluation.  We make the following contributions:
\begin{itemize}
  \item We develop a sparse Bayesian nonparametric  model for SR, learning the number of dictionary elements and the noise variance from the data. 
\item We develop an online variational Bayes (VB) algorithm  finding high quality ``coupled dictionaries"  in a fraction of the time needed by traditional inference.
\item We devise large scale human evaluation experiments to explicitly assess the visual quality of results. 
\end{itemize}
Our approach to SR gives a rich nonparametric representation with scalable learning.

The remainder of the paper is organized as follows: Section \ref{sec:proposed} describes the proposed super-resolution model and non-parametric prior, Section \ref{sec:onlinelearning} contains the derivation of the posterior inference algorithms, Section \ref{base_experiments} presents the experimental results and implementation details, Section \ref{discussion} includes the discussion and future work.

\section{Proposed Approach}
\label{sec:proposed}
Bayesian factor analysis can be used to learn factors / dictionaries from natural images.  Zhou et al. (\cite{zhou2009non}) used  beta process  factor analysis in image denoising, inpainting and compressive sensing.  These models learn both the dictionary elements and their number from the data. 

\begin{figure}
\begin{center}
\includegraphics[scale = 0.1]{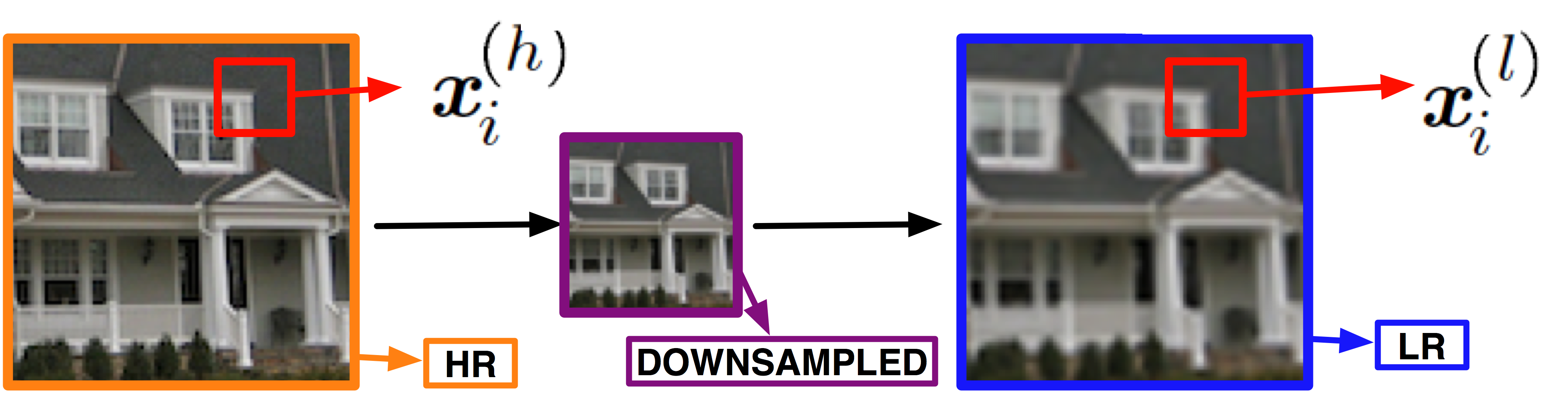}
\caption{\textbf{Depicting the observations extracted} (e.g. image patches)  from high and low resolution images.}
\label{architecture}
\end{center}
\end{figure}

We  build here a nonparametric factor analysis model that couples an HR image to a corresponding LR image. In training, we learn the HR/LR relationship from observed HR/LR pairs. To perform super-resolution, we condition on an observed LR image and compute the conditional expectation of its corresponding HR image. %
A more detailed description of the training process is as follows. We create training data by taking observed HR images and forming corresponding LR images. Figure \ref{architecture} depicts the preprocessing and data extraction steps. We first down-sample the HR images. Then, we up-sample those by interpolating with a deterministic weighting function (e.g. bicubic interpolation). We extract  same-sized patches from the same locations of both the HR and interpolated LR images, and consider those patches as coupled to each other. These are the data on which we train the model. %

In the model, each small patch is generated from  latent global dictionary elements---small images functioning as  factor loadings---using local sparse weights and Gaussian noise. We will first explain how these latent variables are generated and then present how they are used to generate the observations. 

We learn two dictionaries: one for high resolution images and one for low resolution images. In terms of notation, $\textbf{d}_{k}^{(l)}$ represents the LR dictionary element, and $\textbf{d}_{k}^{(h)}$ is the HR dictionary element.  $P^{(l)}$ and $P^{(h)}$  represent the dimensionality of the low and high resolution dictionary elements, respectively. To model  each dictionary element,  we use a zero-mean Gaussian distribution,
\begin{align}
\textbf{d}_{k}^{(l)} \mbox{ } &\thicksim  \mbox{ }  \mathcal{N}(0,P^{(l)^{-1}}\textbf{I}_{P^{(l)}}) & \textbf{d}_{k}^{(h)} \mbox{ } &\thicksim  \mbox{ }  \mathcal{N}(0,P^{(h)^{-1}}\textbf{I}_{P^{(h)}}). \nonumber
\end{align}
The matrix form of the dictionaries are $\textbf{D}^{(l)}$ and $\textbf{D}^{(h)}$ where $k$th columns of those matrices are $\textbf{d}_{k}^{(l)}$ and $\textbf{d}_{k}^{(h)}$, respectively.

Following   \cite{superresolution1}, we assume that the sparse weights are shared by both resolution levels for combining  dictionary elements to produce images. This is the key property of the model that allows us to frame super-resolution as inference. Sparse weights have two components: real valued weights $s_{ik}$ and binary valued assignments $z_{ik}$. To model the weights  $s_{ik}$, we use a zero-mean Gaussian distribution with precision $\alpha$. $\textbf{z}_i$ is a binary vector that encodes which dictionary elements are activated
for the corresponding observation. $p(\textbf{z})$ represents the prior of this variable and we will elaborate on this in next section. These are given as
\begin{align*}
s_{ik} \mbox{ } &\thicksim  \mbox{ }  \mathcal{N}(0,1/\alpha)& z_{ik} \mbox{ } &\thicksim  \mbox{ } p(z_{ik}).\nonumber
\end{align*}

 We place Gamma priors on the precisions of the sparse weights and observation noise ($ \alpha$ and $\gamma$).  The two resolution levels share these variables as well,%
 \begin{align}
\gamma \mbox{ } &\thicksim  \mbox{ } \mathrm{Gamma}(c,d),        &  \alpha &\sim \textrm{Gamma}(e,f) \nonumber.
\end{align}

 Let $\textbf{x}_i^{(h)}$ and $\textbf{x}_i^{(l)}$ represents patches extracted from HR and LR images, respectively, as shown in Figure \ref{architecture}. Given the (global) dictionary elements and (local) sparse weights, the observations are modeled as
\begin{align*}
\boldsymbol{\epsilon}_{i}^{(l)}  &\thicksim    \mathcal{N}(\textbf{0}, \gamma^{-1}\textbf{I}_{P^{(l)}}) &\boldsymbol{\epsilon}_{i}^{(h)}  &\thicksim    \mathcal{N}(\textbf{0}, \gamma^{-1}\textbf{I}_{P^{(h)}}) \\
\textit{\textbf{x}}_i^{(l)}  &=  \textbf{D}^{(l)}(\textbf{s}_i\odot \textbf{z}_i) +  \boldsymbol{\epsilon_i}^{(l)} &\textit{\textbf{x}}_i^{(h)}  &=  \textbf{D}^{(h)}(\textbf{s}_i\odot \textbf{z}_i) +  \boldsymbol{\epsilon_i}^{(h)}\nonumber
\end{align*}
where $\{(l), (h)\}$ represents LR and HR, respectively.  Here, $N$ is the total number of patches, and $\odot$ represents the element-wise multiplication of two vectors. Figure \ref{architecture} illustrates the graphical model.

To use this model in SR, we must be able to compute the
posterior distributions of the hidden variables. In the training phase, we must compute the posterior distributions $p(\textbf{D}^{(h)}, \textbf{D}^{(l)} | \{\textbf{x}_i^{(h)}, \textbf{x}_i^{(l)}\})$  of the dictionaries,  given a collection of HR/LR image pairs.  In testing, we use their posterior expectation to reconstruct  a held-out HR image from an LR image,
\vspace{-0.1cm}
\begin{align}
\mathbb{E} [\textbf{x}_j^{(h)} | \textbf{x}_j^{(l)}, \{\textbf{x}_i^{(h)}, \textbf{x}_i^{(l)}\} ]  \approx \hat{\textbf{D}}^{(h)}(\hat{\textbf{s}}_j\odot \hat{\textbf{z}}_j) \label{construct}
\end{align}
where $\hat{\textbf{D}}^{(h)}$ is the mean of the posterior distribution $p(\textbf{D}^{(h)} | \{\textbf{x}_i^{(h)}, \textbf{x}_i^{(l)}\})$ and $(\hat{\textbf{s}}_j\odot \hat{\textbf{z}}_j)$ are the posterior expectation of the sparse weights from the LR image patches ($\textit{\textbf{x}}_j^{(l)}$) via posterior inference. (We discuss algorithms for posterior inference in Section \ref{sec:onlinelearning}.)

\subsection{Beta-Bernoulli Process Prior (BP)}
\label{sec:BP}
We now discuss the prior for the factor assignments $\textbf{z}_i$.  We use a beta-Bernoulli process (BP) \cite{ICML2011Chen_251,VBforIBP,GhahramaniICA,IBP,paisley2009nonparametric,hbp_ibp,zhou2009non, zhou2011dependent}, a prior on infinite binary matrices which is connected to the Indian buffet process (IBP).  Each row encodes which dictionary elements are activated for the corresponding observation; columns with at least one active cell correspond to factors.  The distinguishing characteristic of this prior is that the number of these factors is not specified a priori. Conditioned on the data, we examine the posterior distribution of the binary matrix to obtain a data-dependent distribution of how many components are needed.

The IBP metaphor gives the intuition. Consider a buffet of dishes at a restaurant. Suppose there are infinite number of dishes and we are trying to specify the infinite binary matrix indicating which customers (observations) choose which dishes (factors/dictionary elements). In the Indian buffet process (IBP), $N$ customers enter the restaurant sequentially. Each customer chooses dishes in a line from a buffet. The first customer starts from the beginning of the buffet and takes  from each dish, stopping after  Poisson($\tau$) number of dishes. The $i$th customer starts from the beginning as well, but decides to take from dishes in proportion to their popularity within the previous $i-1$ customers.  This proportionality can be quantified as $\frac{m_k}{i}$
where $m_k$ is the number of previous customers who took this $k$th dish. After considering the dishes previously taken by other customers, the $i$th customer tries a Poisson($\frac{\tau}{i}$) number of new dishes. Which customers chose which dishes is recorded by the infinite binary matrix with $N$ rows (indicating the customers/observations) and infinite columns (indicating the dishes/factors/dictionary elements). One important (and surprising) property of this process is that the joint probability of final assignment is independent of the order of customers getting into the restaurant which is called exchangeability property of the prior \cite{ibp_JMLR}.

The probabilistic construction is as follows. Each observation $i$ is drawn from a Bernoulli process (a sequence of independent identically distributed Bernoulli trials), $ \textbf{x}_i \sim \mbox{BeP}(B)$ where $B$ is drawn from a beta process $B \sim \mbox{BP}(c_0, B_0)$. $B_0$ represents the base measure with $B_0={\cal N}(0, 1/\beta \textbf{I})$. As $K \rightarrow \infty$, the $i$th observation is $ \textbf{x}_i = \sum_{k=1}^{\infty}z_{ik}\delta_{d_{k}}$ where $z_{ik}$ denotes whether the dictionary element $\textbf{d}_{k}$ is used while representing the $i$th observation or not, and the sample from the beta process is given by $B=\sum_{k=1}^{\infty}\pi_k\delta_{d_{k}}$. Here, $\pi_k$ represents the usage probability of dictionary element $\textbf{d}_k$.

\begin{figure}
\begin{center}
\includegraphics[scale = 0.6]{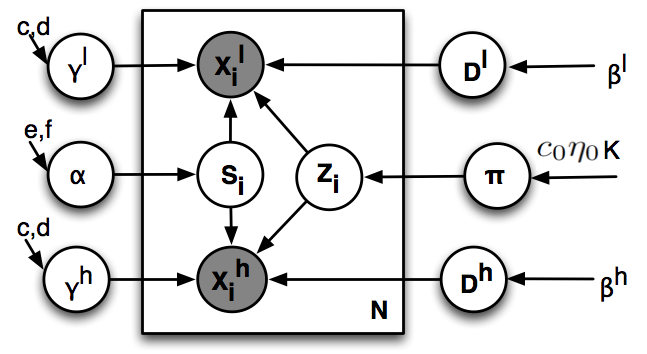}
\end{center}
\caption{\textbf{Graphical Model}.}
\label{baseModel}
\end{figure}

In inference, we use a finite beta-Bernoulli approximation
\cite{IBP}. The finite  model truncates the number of
dictionary elements to $K$ and is given by
\begin{align*}
\pi_k \sim \mathrm{Beta}(c_0\eta_0, c_0(1-\eta_0)), \mbox{ }z_{ik} \sim \textrm{Bernoulli}(\pi_k) 
\end{align*}
where $c_0$ and $\eta_0$ are
scalars and $\quad k \in 1, \ldots K$.  %
As $K$ tends to infinity, the finite beta-Bernoulli approximation approaches the IBP/BP.  If the truncation is large enough, data
analyzed with this prior will exhibit fewer than $K$
components~\cite{VBforIBP}.

\subsection{Super-resolution via Posterior Distributions}
\label{inference}

Our algorithm has 2 stages: fitting the model on pairs of HR and LR images, and super-resolving new LR images to create HR versions.

\textbf{Training: Coupled Dictionary Learning Stage.}
In training, we observe $\textbf{x}_i^{(h)}$ and $\textbf{x}_i^{(l)}$.  All other random variables are latent.   The key inference problem to be solved is the computation of the posterior distributions of the hidden variables. In the training phase, we must compute the posterior distributions $p(\textbf{D}^{(h)}, \textbf{D}^{(l)} | \{\textbf{x}_i^{(h)}, \textbf{x}_i^{(l)}\})$  of the dictionaries  given a collection of HR/LR image pairs. We rewrite the coupled model in a form similar to the single scale model:
\begin{align}
  \textit{\textbf{x}}_i^{(c)} =
 \!\left(\!
    \begin{array}{c}
       \textit{\textbf{x}}_i^{(l)} \\
      \textit{\textbf{x}}_i^{(h)}
    \end{array}
  \!\right)\!,
  \textbf{d}_{k}^{(c)} = 
 \! \left(\!
    \begin{array}{c}
       \textbf{d}_{k}^{(l)} \\
      \textbf{d}_{k}^{(h)}
    \end{array}
  \!\right)\!,
\boldsymbol{\epsilon}_{i}^{(c)} = 
  \!\left(\!
    \begin{array}{c}
\boldsymbol{\epsilon}_{i}^{(l)} \\
\boldsymbol{\epsilon}_{i}^{(h)}
    \end{array}
  \!\right)\!  
  \label{eq:coupling}
\end{align}
where the superscript $(c)$ corresponds to combination of $(l)$ and $(h)$.  Writing the fully-observed model in this way reveals that we can train the dictionaries with similar methods as for the single-scale base model (Training amounts to approximating the posteriors of these values).  The differences are that we use combined patches $\textit{\textbf{x}}_i^{(c)}$ and combined dictionaries $\textbf{d}_{k}^{(c)}$.  This leads to shared sparse weights for the two resolution levels. (The details of how we compute the distribution $p(\textbf{D}^{(h)}, \textbf{D}^{(l)} | \{\textbf{x}_i^{(h)}, \textbf{x}_i^{(l)}\})$ are discussed in Section \ref{sec:onlinelearning}.)

\textbf{Super-resolving a Low Resolution Image.}  With fitted dictionaries in hand, we now show how to form HR images from LR images via posterior computation.

In this prediction setting, the HR image $\textbf{x}_i^{(h)}$ is unknown; the goal is  to reconstruct it from the LR image patches $\textbf{x}_i^{(l)}$, the posterior estimates of the dictionaries $(\hat{\textbf{D}}^{(h)}, \hat{\textbf{D}}^{(l)})$, and the precisions $\hat{\gamma}, \hat{\alpha}$ of the noise and the sparse weights,   %
\begin{align*}
  \textit{\textbf{x}}_i^{(c)} =
  \left(
    \begin{array}{c}
       \textit{\textbf{x}}_i^{(l)} \\
     -
    \end{array}
  \right),\mbox{ }
  \textbf{d}_{k}^{(c)} = 
  \left(
    \begin{array}{c}
       \textbf{d}_{k}^{(l)} \\
      \textbf{d}_{k}^{(h)}
    \end{array}
  \right),\mbox{ }
\boldsymbol{\epsilon}_{i}^{(c)} = 
  \left(
    \begin{array}{c}
\boldsymbol{\epsilon}_{i}^{(l)} \\
-
    \end{array}
  \right).
\end{align*}
First we find estimates of the sparse factor scores, $(\hat{\textbf{s}}_i\odot \hat{\textbf{z}}_i)$, by using the LR image patches $\textit{\textbf{x}}_i^{(l)}$ and posterior estimates of the dictionaries and precisions $\gamma$ and $\alpha$. The fitted value of  $\alpha$ determines the strength of a ``regularization term" that controls the sparsity of the factor scores.

More precisely,  this prediction setting has 3 steps. The input is a set of held-out LR image patches $\textbf{x}_i^{(l)}$, the posterior estimates of the dictionaries $(\hat{\textbf{D}}^{(h)}, \hat{\textbf{D}}^{(l)})$, and the precisions $\hat{\gamma}, \hat{\alpha}$ of the noise and the sparse weights. The steps are as follows:  %
\begin{enumerate}
\item We find estimates of the sparse factor scores, $(\hat{\textbf{s}}_i\odot \hat{\textbf{z}}_i)$, conditioned on the LR image patches $\textit{\textbf{x}}_i^{(l)}$ and estimates $(\hat{\textbf{D}}^{(h)}, \hat{\textbf{D}}^{(l)}), \hat{\gamma}, \hat{\alpha}$ from the training stage.  
\item Eq. \ref{construct} determines the HR patches $ \hat{\textbf{x}}_i^{(h)}\!$. 
\item We replace each $\textbf{x}_i^{(l)}$ by its corresponding collocated $\hat{\textbf{x}}_i^{(h)}\!$;  the whole HR image, $ \hat{\textbf{X}}^{(h)}\!$, is the pixel-wise average of those overlapping reconstructions.
\end{enumerate}

  \textbf{Post-processing:} Following \cite{superresolution1}, we  apply a post-processing step that, when down-sampled, the reconstructed HR image, $ \hat{\textbf{X}}^{(h)}$, should match the given LR image $ \textbf{X}^{(l)}$. Specifically, we solve the following optimization:
  \begin{align}
  \hat{\textbf{X}}^{(h)*} = \underset{\textbf{X}}{\operatorname{argmin}} || f(\textbf{X}) -  \textbf{X}^{(l)} ||_{2}^{2} + c|| f(\textbf{X}) -  \hat{\textbf{X}}^{(h)} ||_{2}^{2} 
     \nonumber
\end{align}
where  $f()$ is a linear operator consisting of an anti-aliasing filter followed by down-sampling. This optimization problem is solved with gradient descent. %

\section{Posterior Inference}
\label{sec:onlinelearning}

In the proposed approach, all of the priors are in the conjugate
exponential family. In a first implementation, we use Gibbs sampling. We iteratively sample from the conditional distribution of each hidden variable given the others and the observations. This defines a Markov chain whose stationary distribution is the posterior \cite{robert_casella}. The corresponding sampling equations are analytic and provided in the
appendix A-B (appendix is in the supplementary material).

The Gibbs sampler has difficulty with scaling to large data, because
it must go through many iterations, each time visiting the entire data
set before the sampler mixes. For this reason, both our Gibbs sampler
and ScSR use $10^5$ patches sampled from $3\times10^6$. We now develop
here an alternative algorithm to Gibbs sampling for SR that scales to large and
streaming data. Specifically, we develop an  online variational inference algorithm.

Variational inference is a deterministic alternative to MCMC that replaces
sampling with optimization. The idea is to posit a parameterized
family of distribution over the hidden variables and then optimize the
parameters to minimize the KL divergence to the posterior of interest
\cite{VBintro}. Our algorithm iteratively tracks an approximate
posterior distribution, which improves as more data are seen.

In typical applications, the variational objective is optimized with
coordinate ascent, iteratively optimizing each parameter while holding the others fixed. However, in Bayesian settings, this suffers from
the same problem as Gibbs sampling---the entire data set must be swept
through multiple times in order to find a good approximate posterior.
In the algorithm we present here, we replace coordinate ascent
optimization with \textit{stochastic optimization}---at each
iteration, we subsample our data and then adjust the parameters
according to a noisy estimate of the gradient. Because we only to
subsample the data at each iteration, rather than analyze the whole
data set, the resulting algorithm scales well to large data. This
technique was pioneered in~\cite{Sato} and was recently exploited for
online learning of topic models~\cite{onlineLDA} and hierarchical
Dirichlet processes~\cite{onlineHDP}.

We first develop the coordinate ascent algorithm for the coupled model. Then we
derive the online variational inference algorithm, which can more
easily handle large data sets.

\subsection{Variational Inference for the Coupled model}

\label{VB}
We use the coupling perspective in Section \ref{inference} to derive  the batch variational Bayes (VB) algorithm.  The single-scale base model  is the BPFA model of \cite{paisley2009nonparametric}, which  gives a mean-field variational inference algorithm. The batch VB algorithm derived here is the coupled version of that.

We first define  a parametrized
family of distributions over the hidden variables.  Let $\textbf{Q} =
\{\boldsymbol{\pi}, \textbf{Z, S, D}, \gamma, \alpha \}$
denote the hidden variables for all $i, k$. We write coupled data as in Equation \ref{eq:coupling}; in the new set-up the variables to be learned become $\textbf{Q} =\{\boldsymbol{\pi}, \textbf{Z, S}, \textbf{D}^{(c)}, \gamma, \alpha \}$.  We use a fully
factorized variational distribution,
\begin{equation*}
  q(\textbf{Q}) =
  q_{\tau}(\boldsymbol{\pi})q_{\phi}(\textbf{D}^{(c)})q_{\nu}(\textbf{Z})q_{\theta}(\textbf{S})q_{\lambda}(\boldsymbol{\gamma})q_{\epsilon}(\boldsymbol{\alpha}).
  \end{equation*}
Each component of this distribution is governed by a free variational
parameter,
\begin{align*}
q_{\tau_k}(\pi_k) &= \mathrm{Beta}( \tau_{k1}, \tau_{k2}) & q_{\nu_{ik}}(z_{ik}) &= \mathrm{Bernoulli}( \nu_{ik})\nonumber\\
q_{\phi_{kj}}(d_{kj}) &= \mathcal{N}(\phi_{kj}, \Phi_{kj}) & q_{\lambda}(\gamma) &= \mathrm{Gamma}(\lambda_{1}, \lambda_{2})\nonumber\\
 q_{\theta_{ik}}(s_{ik}) &= \mathcal{N}(\theta_{ik}, \Theta_{ik})&q_{\epsilon}(\alpha) &= \mathrm{Gamma}(\epsilon_{1}, \epsilon_{2}) \nonumber
\end{align*}
We optimize these parameters with respect to a bound on
the marginal probability of the observations.  This bound is
equivalent, up to a constant, to the negative KL divergence between
$q$ and the true posterior. Thus maximizing the bound is equivalent to
minimizing KL divergence to the true posterior. Let $\boldsymbol{\Xi}
= \{c_0,\eta_0,c,d,e,f\}$ be the hyper-parameters.  The variational lower
bound is
\begin{align}
 & \log(p(\textbf{X}^{(c)}|\boldsymbol{\Xi})) \geq  H(q)+ \sum\nolimits_{k
    =1}^{K}\Big\{\mathbb{E}_{\textbf{q}}[\log p(\pi_k|c_0, \eta_0, K)] \nonumber\\
    &+
  \sum\nolimits_{i =
    1}^{N}\mathbb{E}_{\textbf{q}}[\log
  p(z_{ik}|\boldsymbol{\pi})]
 +\sum\nolimits_{j=1}^{J}\mathbb{E}_{\textbf{q}}[\log\big(p(d_{kj}|\beta_{kj})
  \big)] \nonumber \\ 
  &   + \sum\nolimits_{i = 1}^{N}\mathbb{E}_{\textbf{q}}[\log \big( p(s_{ik}|\alpha)  p(\alpha|e,f) \big)] \Big\} \label{elbo}\\
  &+\sum\nolimits_{i = 1}^{N}\Big\{\mathbb{E}_{\textbf{q}}[\log
  p(\textbf{x}_{i}^{(c)}|\textbf{Z,S,}\textbf{D}^{(c)}\gamma)]+\mathbb{E}_{\textbf{q}}[\log p(\gamma|c,d)]\Big\}, \nonumber
\end{align}
where $\textrm{H}(q)$ is the entropy of the variational distribution and dimensionality of the dictionary elements $J$ is twice as big as the single-scale model.
We denote this function ${\cal L}(q)$.

Holding the other parameters fixed, we can optimize each variational
parameter exactly; this gives an algorithm that goes uphill in
${\cal L}(q)$~\cite{bishop}.  (Further, this will provide the
algorithmic components needed for the online algorithm of
Section~\ref{sec:onlinevb}.)

 Update equations for each free parameter optimizing this bound are given below. In all equations, $\textbf{I}_P$ represents $P\times P$ identity matrix, and $\tilde{\textbf{x}}_{i(-k)}^{(c)} $ represents the reconstruction error using all but  the  $k$th dictionary element, that is   
\[\tilde{\textbf{x}}_{i(-k)}^{(c)} = \textbf{x}_{i}^{(c)} - \textbf{D}^{(c)}(\textbf{s}_i\odot \textbf{z}_i) + \textbf{d}_k^{(c)}(s_{ik}\odot z_{ik}).\]
The expectation based on the variational distribution is then given by
\[ \mathbb{E}_q[\tilde{\textbf{x}}_{i(-k)}^{(c)}] = \textbf{x}_{i}^{(c)}  + \boldsymbol{\phi}_{k}^{(c)}(\theta_{ik} \nu_{ik}) - \sum_{k=1}^{K}\boldsymbol{\phi}_{k}^{(c)}(\theta_{ik} \nu_{ik}).\]

\textbf{Update for the binary factor assignment $z_{ik}$}: The variational parameter for factor assignment $z_{ik}$ is $ \nu_{ik}$. We first consider two values of the variational distribution for two values (0,1) of $z_{ik}$, 
\begin{align*}
q(z_{ik}  = 1) \mbox{  } &\propto \mbox{  }  \exp( \mathbb{E}_q[\ln(\pi_k)]) \mbox{exp}\Big( -  \frac{  \frac{\lambda_1}{\lambda_2} \big( ( \theta_{ik}^2+\Theta_{ik})  (\boldsymbol{\phi}_{k}^{(c)^T} \boldsymbol{\phi}_{k}^{(c)} + \sum_{j}\Phi_{kj}) - 2\theta_{ik} \boldsymbol{\phi}_{k}^{(c)^T}\mathbb{E}_q[\tilde{\textbf{x}}_{i(-k)}^{(c)}]  \big)}{2}\Big)  \\
q(z_{ik}  = 0) \mbox{  } &\propto \mbox{  }  \exp(\mathbb{E}_q[ln(1-\pi_{k})]), \mbox{   }\mbox{where}
\end{align*}
\begin{align*}
 \mathbb{E}_q[\ln(\pi_k)]  \mbox{  } &= \mbox{  }   \psi\big(c_0\eta_0 + \sum_{i}\nu_{ik}\big) -  \psi(c_0 + N)\\
 \mathbb{E}_q[\ln(1-\pi_k)]  \mbox{  } &= \mbox{  }   \psi\big(c_0(1-\eta_0) - \sum_{i}\nu_{ik} + N\big) -  \psi(c_0 + N) 
 \end{align*}
  Then the update equation for the variational parameter  $ \nu_{ik}$ is given as
 \begin{align*}
 \nu_{ik}   \mbox{  } &= \mbox{  }   \frac{q(z_{ik}  = 1|-)}{q(z_{ik}  = 1|-)+q(z_{ik}  =01|-)}
\end{align*}

\textbf{Update for the shared sparse weight $s_{ik}$}: The variational distribution for the sparse weight $s_{ik}$ is Gaussian parametrized with mean $\theta_{ik} $ and variance $\Theta_{ik} $. Coordinate ascent update equation for these free variational parameters are
\begin{align*}
\Theta_{ik}  \mbox{ } &= \mbox{ } \Big( \frac{\epsilon_1}{\epsilon_2} + \frac{\lambda_1}{\lambda_2}\nu_{ik} (\boldsymbol{\phi}_{k}^{(c)^T} \boldsymbol{\phi}_{k}^{(c)} + \sum_{j}\Phi_{kj}) \Big)^{-1}\\
\theta_{ik}  \mbox{ } &= \mbox{ } \frac{\lambda_1}{\lambda_2} \Theta_{ik}\nu_{ik}  \boldsymbol{\phi}_{k}^{(c)^T}\mathbb{E}_q[\tilde{\textbf{x}}_{i(-k)}^{(c)}] .
\end{align*}

\textbf{Update for the $k$th coupled dictionary element } $ \textbf{d}_{k}^{(c)} $:  The variational distribution for the couple dictionary element $\textbf{d}_{k}^{(c)} $ is Gaussian parametrized with mean $\boldsymbol{\phi}_{k}^{(c)} $ and variance $\boldsymbol{\Phi}_{k}^{(c)}  $. Coordinate ascent update equation for these free variational parameters are

\begin{align*}
\boldsymbol{\Phi}_{k}^{(c)}  \mbox{ } &= \mbox{ } \Big( 2P \textbf{I}_{2P}  + \frac{\lambda_1}{\lambda_2}\sum_{i=1}^{N}( \theta_{ik}^2+\Theta_{ik})  \nu_{ik} ^2\Big)^{-1}\\
\boldsymbol{\phi}_{k}^{(c)}  \mbox{ } &= \mbox{  }  \frac{\lambda_1}{\lambda_2}\boldsymbol{\Phi}_{k} ^{(c)}  \sum_{i=1}^{N}\theta_{ik}  \nu_{ik} \mathbb{E}_q[\tilde{\textbf{x}}_{i(-k)}^{(c)}]. 
\end{align*}
 The updates for high resolution (h) and low resolution (l) components can be given separately as 
\begin{align*}
\boldsymbol{\Phi}_{k}^{(h)}  \mbox{ } &= \mbox{ }  \Big( 2P \textbf{I}_{P}  + \frac{\lambda_1}{\lambda_2}\sum_{i=1}^{N}( \theta_{ik}^2+\Theta_{ik})  \nu_{ik} ^2\Big)^{-1} 
&\boldsymbol{\Phi}_{k}^{(l)}  \mbox{ } &= \mbox{ } \Big( 2P \textbf{I}_{P}  + \frac{\lambda_1}{\lambda_2}\sum_{i=1}^{N}( \theta_{ik}^2+\Theta_{ik})  \nu_{ik} ^2\Big)^{-1}\\
\boldsymbol{\phi}_{k}^{(h)}  \mbox{ } &= \mbox{  }  \frac{\lambda_1}{\lambda_2}\boldsymbol{\Phi}_{k}^{(h)}   \sum_{i=1}^{N}\theta_{ik}  \nu_{ik}  \tilde{\textbf{x}}_{i(-k)}^{(h)}  
&\boldsymbol{\phi}_{k}^{(l)}  \mbox{ } &= \mbox{  }  \frac{\lambda_1}{\lambda_2}\boldsymbol{\Phi}_{k}^{(l)}   \sum_{i=1}^{N}\theta_{ik}  \nu_{ik}  \tilde{\textbf{x}}_{i(-k)}^{(l)}. 
\end{align*}

\textbf{Update for  the dictionary usage probabilities $\pi_k$}: The variational distribution for the dictionary usage probabilities $\pi_k$ is a beta distribution parametrized with the shape parameters ($\tau_{k1}, \tau_{k2}$). Coordinate ascent update equation for these free variational parameters are
\begin{align*}
\tau_{k1}  \mbox{  } &= \mbox{ } c_0 \eta_0+  \sum_{i=1}^{N}\nu_{ik}  \\
\tau_{k2}   \mbox{  } &= \mbox{ }  \mbox{ } N-  \sum_{i=1}^{N}\nu_{ik} + c_0(1-\eta_0).
\end{align*}

\textbf{Update for  the precision $\gamma$}: The variational distribution for the precision $\gamma$ of the observation noise $\boldsymbol{\epsilon}_i$ is a gamma distribution parametrized with ($\lambda_1, \lambda_2$). Coordinate ascent update equation for these free variational parameters are
\begin{align*}
\lambda_1  \mbox{ } &= \mbox{  }  c + NP\\
\lambda_2  \mbox{ } &= \mbox{  }  d + \frac{1}{2}\sum_{i=1}^{N} \Big \{ ||\textbf{x}_{i}^{(c)}  - \sum_{k=1}^{K}\boldsymbol{\phi}_{k}^{(c)}(\theta_{ik} \nu_{ik}) ||_2^2 +\sum_{k=1}^{K}\nu_{ik}( \theta_{ik}^2+\Theta_{ik})  (\boldsymbol{\phi}_{k}^{(c)^T} \boldsymbol{\phi}_{k}^{(c)} + \sum_{j}\Phi_{kj})\\
\mbox{ } &- \mbox{  } \sum_{k=1}^{K} \nu_{ik}  \boldsymbol{\phi}_{k}^{(c)^T} \boldsymbol{\phi}_{k}^{(c)}   \theta_{ik}^2 \Big \}.
\end{align*}

\textbf{Update for  the precision $\alpha$}: The variational distribution for the precision $\alpha$ of the sparse weights $s_{ik}$ is a gamma distribution parametrized with ($\epsilon_1, \epsilon_2$). Coordinate ascent update equation for these free variational parameters are
\begin{align*}
\epsilon_1 \mbox{ } &= \mbox{  }   e + \frac{1}{2}NK\\
\epsilon_2  \mbox{ } &= \mbox{  }    f + \frac{1}{2}\sum_{i=1}^{N} \sum_{k=1}^{K}(\theta_{ik}^2+\Theta_{ik}^2).
\end{align*} 

  \begin{algorithm} [h!]
\caption{Batch VB} 
\label{alg2} 
\begin{algorithmic} 
\small
\STATE Sample $N$ observations from the data. Initialize $\boldsymbol{\tau}, \boldsymbol{\nu}, \boldsymbol{\phi}, \boldsymbol{\Phi}, \boldsymbol{\theta},  \boldsymbol{\Theta}, \boldsymbol{\lambda}, \boldsymbol{\epsilon}$ using Gibbs sampler. 
\FOR{$t=1$ to $T$}
\STATE Init. local variables $\nu_{n_k}, \boldsymbol{\theta}_{n_k},  \boldsymbol{\Theta}_{n_k}$ using Gibbs sampler.
\WHILE{relative improvement in $\ell$ is large } 
\FOR{$k = 1$ to $K$}
\FOR{$n = 1$ to $N$}
\STATE update $\nu_{nk}, \boldsymbol{\theta}_{nk},  \boldsymbol{\Theta}_{nk}$ by using batch VB updates. 
\ENDFOR
\STATE compute  $\Phi_{k}, \boldsymbol{\phi}_{k},  \boldsymbol{\tau}_{k}, \boldsymbol{\lambda}, \boldsymbol{\epsilon}$ by batch VB updates.
\ENDFOR
\ENDWHILE
\ENDFOR
\end{algorithmic}
\end{algorithm}

\subsection{Online Variational Inference}
\label{sec:onlinevb}
We now develop online variational inference.  We divide the variational parameters into \textit{global} variables
and \textit{local} variables.  Global variables depend on all of the
images.  These are the dictionary probabilities $\boldsymbol{\pi}_k$,
dictionary elements $\textbf{d}_k$, precisions $\boldsymbol{\alpha}$ and
$\gamma$.  Local variables are the ones drawn for each image.  These are the weights $\textbf{s}_{i}$,
binary variables $\textbf{z}_{i}$. The algorithm  iterates between
optimizing the local variables using local (per-image) coordinate
ascent, and optimizing the global variables.  This same structure is found in many Bayesian nonparametric models ~\cite{VBforIBP, HDP}.
  
   The basic idea is to optimize Equation~\ref{elbo}
via stochastic optimization~\cite{Robbins:1951}. This means we repeatedly follow noisy estimates of the gradient
with decreasing step sizes $\rho_t$.  If the step sizes satisfy
$\textstyle \sum_t \rho_t = \infty$ and $\textstyle \sum_t \rho^2_t
<\infty$ then we will converge to the optimum of the objective.  (In
variational inference, we will converge to a local optimum.)

  \begin{algorithm} [h!]
\caption{Online VB with mini-batches} 
\label{alg3} 
\begin{algorithmic} 
\small
\STATE Define $\rho_t = (r+t)^{-\kappa}$,  Initialize $\boldsymbol{\tau}, \boldsymbol{\nu}, \boldsymbol{\phi}, \boldsymbol{\Phi}, \boldsymbol{\theta},  \boldsymbol{\Theta}, \boldsymbol{\lambda}, \boldsymbol{\epsilon}$ using Gibbs sampler. 
\FOR{$t=1$ to $\frac{N}{N_S}$}
\STATE Sample $N_S$ new observations from the data. Initialize local variables $\nu_{n_k}, \boldsymbol{\theta}_{n_k},  \boldsymbol{\Theta}_{n_k}$ using Gibbs sampler.
\WHILE{relative improvement in $\ell$ is large } 
\FOR{$k = 1$ to $K$}
\FOR{$n_t = (t-1) \times N_S + 1$ to $t \times N_S $}
\STATE update $\nu_{n_tk}, \boldsymbol{\theta}_{n_tk},  \boldsymbol{\Theta}_{n_tk}$ by using batch VB updates. 
\ENDFOR
\STATE compute  $\tilde{\Phi}_{k}, \tilde{\boldsymbol{\phi}}_{k},  \tilde{\boldsymbol{\tau}}_{k}, \tilde{\boldsymbol{\lambda}}, \tilde{\boldsymbol{\epsilon}}$ by batch VB updates as if there are $N /N_S$ copies of the images.
\ENDFOR
\ENDWHILE
\FOR{$k = 1$ to $K$}
\STATE update $\Phi_{k}, \boldsymbol{\phi}_{k}, \boldsymbol{\tau}_{k},  \boldsymbol{\lambda}, \boldsymbol{\epsilon}$ by  Equation \ref{up4}
\ENDFOR
\ENDFOR
\end{algorithmic}
\end{algorithm} 

The noisy estimates of the gradient are obtained from subsampled data. We
write the objective $\mathcal{L}$ as a sum over  data points. Defining the distribution $g(n)$ which uniformly samples from the data, we can then write $\mathcal{L}$ as an expectation under this distribution,
\begin{align}
  \mathcal{L} &=\sum\nolimits_{n=1}^{N}\ell(\boldsymbol{\tau},
  \boldsymbol{\nu}_n, \boldsymbol{\phi},
  \boldsymbol{\theta}_n, 
    \boldsymbol{\lambda}_n, \boldsymbol{\epsilon}_n, \textbf{X}_n).
  \label{subobjective}\\
  &=N\mathbb{E}_{g}[\ell(\boldsymbol{\tau},
\boldsymbol{\nu}_n, \boldsymbol{\phi}, 
\boldsymbol{\theta}_n, \boldsymbol{\Theta}_n, \boldsymbol{\lambda}_n,
\boldsymbol{\epsilon}_n,  \textbf{X}_n)
]\label{newobjective}
\end{align}
The gradient of the objective can be written as a similar expectation. Thus, sampling data at random and computing
the gradient of $\ell_n$ gives a noisy estimate of the gradient.  

\begin{figure}
\begin{center}
\subfigure{\includegraphics[scale = 0.9]{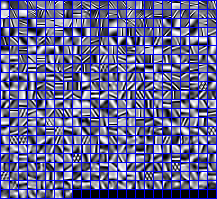}}
\subfigure{\includegraphics[scale = 0.9]{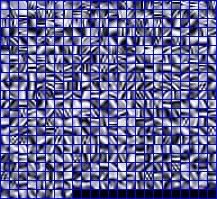}}
\caption{\textbf{Dictionary trained in batch mode} on lluminance channel with SR ratio = 2. (Left) HR Dictionary, (Right) LR Dictionary, Every square represents a dictionary element and the HR-LR pairs  are co-located. HR dictionary consists of  sharper edges.}
\label{dictionaryGaussianPoisson2}
\end{center}
\end{figure}

There are two further simplifications.  First, when we subsample the
data we optimize the local variational parameters fully and
compute the gradient of $\ell_n$ with respect to only the global
variational parameters.  Second, we use the natural
gradient~\cite{amari} rather than the gradient.  In mean field
variational inference, this simplifies the gradient step as follows.
Suppose we have sampled an image $n$ and fitted its local variational
parameters given the current settings of the global variational
parameters.  Let $\tilde{\boldsymbol{\tau}},
\tilde{\boldsymbol{\phi}}, \tilde{\boldsymbol{\Phi}}, \tilde{\boldsymbol{\lambda}}, \tilde{\boldsymbol{\epsilon}}$ be the global variational updates from
Section \ref{VB} as though we observed $N$ copies of
that image.  (Note that these depend on its local variational
parameters.)  Following a noisy estimate of the natural gradient of
${\cal L}$ is equivalent to taking a weighted average of the current
and the newly fitted global parameters, e.g. 
\begin{align}
  \boldsymbol{\phi} = (1-\rho_t)\boldsymbol{\phi} + \rho_t\tilde{\boldsymbol{\phi}}.  \label{up4}
\end{align}
It follows that  there is no additional
computational cost to optimizing the global variational parameters
with stochastic optimization versus coordinate ascent.

In our implementation, we decrease the step-size $\rho_t$ by $\rho_t = (\rho_0 +
t)^{-\kappa}$. The learning rate parameter $\rho_0$ down-weights early
iterations; the parameter $\kappa$ controls the speed of forgetting
previous values of the global variables.

The full online VB algorithm is listed in Algorithm \ref{alg3}.  (Note
that we sample the data in mini-batches, rather than one at a
time.  When the mini-batch size is equal to one data point, we recover
the algorithm as described above.)

\subsection{Initialization with MCMC.} 
We initialize both the batch and
online VB with a few iterations (e.g. 5 or 10) of MCMC.\footnote{For batch VB, these MCMC samples are collected on the same subset of the data on which batch VB will process. For online VB, they are collected from the mini-batches. In both cases, scale problem of MCMC is not an issue since we only collect few samples (e.g. 5 or 10). As we mentioned before, scale is a problem for MCMC since it needs to go over the data many times for convergence (e.g. thousands of iterations). Time scaling is discussed in more detail in Section \ref{section:time} } This is useful for two
reasons: (1) It provides  a good initialization and thus faster
convergence, (2) Noisy random-walks of MCMC help VB avoid low-quality local optima:  at the beginning of each e-step, MCMC initializes $\textbf{s}_i$ and $\textbf{z}_i$ by sampling from their approximate posterior distribution, given the most recent global variables. These samples are noisy estimates of the sparse weights near their posterior means. For instance, when the factor assignment $z_{ik}$ equals $0$, the MCMC draws the sparse weight $s_{ik}$ from  the prior $ \mathcal{N}(0, 1/\alpha)$ whereas in VB it would be exactly 0. Providing the freedom to ``jiggle" gives the algorithm the opportunity, similar to simulated annealing, to jump away from one local optimum to reach a better optimum.

\section{Experiments}
\label{base_experiments}

We use three data sets.  To train, we use the set of 68 images collected from the web by \cite{superresolution1}.  We test on the Berkeley natural image data set (20 100$\times$100 images) and a benchmark set of images (11 images of various size) used by the community to evaluate SR algorithms.\footnote{We are using SR ratio=2 or 4. For SR ratio 2, the images which do not have even number of rows/columns are cut to have even number of rows/column to prevent any possible mismatch and error in computing PSNR in all algorithms. For instance the last column of pixels from an image of size $330 \times 171$ is excluded so the corresponding image have the size $330 \times 170$. 
} These data sets provide us with a rich set of HR-LR pairs.

  \begin{table}
       \caption{Test Results with SR ratio = 2. PSNR for the illuminance channel is presented (the higher the better). \textbf{BP}: Proposed algorithm trained via Gibbs sampler, \textbf{O-BP}  Proposed algorithm  trained via Online VB, seeing more data,   \textbf{ScSR}: Super-Resolution via Sparse Representation   \cite{superresolution1}, \textbf{NNI}: Nearest neighbor interpolation, \textbf{SME}: Sparse Mixing Estimation \cite{SME}. }
   \begin{center}   
  \begin{tabular}{| c | c | c | c | c |c |c |c |c |}
    \hline
 \textbf{PSNR}		& Bic.      & NNI      & Bil. & SME        & ScSR256  &ScSR512   & BP    & OBP  \\\hline
Baboon         &  23.63  &  23.12  & 23.05 & 23.10  & 24.33     &  24.36  &   24.27  &  \textbf{24.39}\\\hline
Barbara        &  25.35  &  25.10  & 24.92 & 24.42  & 25.88     &  25.89  &    25.98 &  \textbf{25.99}\\\hline
Boat         &  29.95  &  28.39  &  28.94 & 29.72  & 31.23   &  31.29  &    31.17  &  \textbf{31.31}\\\hline
Camera        &  30.32  &  \textbf{35.20}  &  28.94  & 26.33 & 30.68     &  30.46  &   31.51  &  30.94\\\hline
House         &  32.79  &  30.34  &  31.61 & 33.28 &  34.26     &  \textbf{34.31}  &    34.08 &  34.27\\\hline
Peppers         &  31.99  &  29.88  &  31.18 & 33.06 & 33.05    &  33.06  &    32.45  &  \textbf{33.08}\\\hline
Parthen.	        &  28.12  &  27.28  &  27.42 & 27.28 & 29.05     &  \textbf{29.10}  &  28.96  &  29.06\\\hline
Girl  	        &  34.76  &  33.44  & 33.98 & 33.98  & 35.57  &  35.58  &  35.62  &  \textbf{35.66}\\\hline
Flower       &  40.04  &  37.96  & 38.94 & 39.72   &  41.06   &  41.11  &  41.26  &  \textbf{41.33}\\\hline
Lena        &  32.83  &  31.00  &  31.72 & 33.57  & 34.47     &  34.54  &  34.56  &  \textbf{34.68}\\\hline
Raccoon         &  30.95  &  29.82  &  29.95 & 31.73  & 32.39    &  32.43  &  32.43  &  \textbf{32.62}\\\hline
  \end{tabular}     
   \label{tab:benchresults}
    \end{center}
 \end{table}
 
Throughout this work, unless otherwise mentioned we use the same parameters (without any tuning): we set the SR ratio  to 2 or 4   and the patch size to $8\times 8$.\footnote{The visual results for  SR ratio 4 are in the appendix G.} The hyper-parameters are $c=d=e=f=10^{-6}$ and $c_0=2, \eta_0=0.5$, these are standard uninformative priors used in e.g. \cite{zhou2009non}. The truncation level $K$ in BP is set to 512. Most images use fewer factors, e.g. Baboon uses 487, House 438 and Barbara 471 factors.  We apply all algorithms only to the illuminance channel and use Bicubic interpolation for the color layers (Cb, Cr) for all compared methods.

We study our methods with two kinds of posterior inference---Gibbs
sampling (BP) and online variational inference (O-BP), which scales to
larger data sets.\footnote{The software for each algorithm presented and all of the visual results will be
publicly available.} To compare, we study both interpolation and
example-based algorithms. Bicubic interpolation is the gold standard
in the SR literature. We also study nearest neighbor interpolation, bilinear interpolation and sparse mixing estimation (SME) \cite{SME}.  To compare with an example-based method,
we use super-resolution via sparse representation (ScSR,
\cite{superresolution1}). \footnote{We used the code and implementation provided by  \cite{superresolution1}. We also used their training images, in order to have a fair comparison, and we did not change any of their parameters (including noise variance).}   \footnote{We provide visual comparisons to \cite{freemanSR,kimSR, glasnerSR,fattalSR} in appendix H. \cite{glasnerSR}  provides very sharp edges by artificially enhancing them. However, this makes  images  unrealistic (looking like graphically rendered).  Sparse coding allows any single-image SR algorithm as a pre-processing step. Instead of bicubic interpolation (see Figure \ref{architecture}) \cite{glasnerSR}  might be used with sparse coding to boost the sharpness of the edges. } 
\footnote{ The dependent hierarchical Beta process (dHBP),  another bayesian nonparametric prior,  is proposed in \cite{zhou2011dependent}.  It removes the exchangeability assumption of beta-Bernoulli construction. This prior assumes that each observation $i$ has a covariate $\ell_i \in \mathbb{R}^{\mathcal{L}}$.  In this model, the closer the two sparse factor assignments $z_i$ and $z_j$ in the covariate space, the more likely they share similar dictionary elements.  It applies dHBP using spatial information as covariates to image inpainting and spiky noise removal, and shows significant improvement over BP.  We obtained  preliminary results with dHBP for super-resolution. However, in this setting we did not observe improvement over BP.}
Both BP's and ScSR's dictionary
learning stages use $10^5$ patches sampled
from the training data, however O-BP uses the whole set in online fashion. The HR and LR dictionaries trained by our approach
are shown in Figure \ref{dictionaryGaussianPoisson2}. The HR dictionary consists of  sharper edges.

 \begin{table}
       \caption{Test Results with SR ratio = 2. PSNR for the illuminance channel is presented (the higher the better). \textbf{BP}: Proposed algorithm trained via Gibbs sampler, \textbf{O-BP}  Proposed algorithm  trained via Online VB, seeing more data,   \textbf{ScSR}: Super-Resolution via Sparse Representation   \cite{superresolution1}, \textbf{NNI}: Nearest neighbor interpolation. }
   \begin{center}   
  \begin{tabular}{| c | c | c | c | c |c |c |c |}
    \hline
  \textbf{PSNR} & Bic.  & NNI & Bilin. & ScSR256  &ScSR512     & BP    & O-BP  \\\hline
N1  &  29.74  &  27.44  &  28.39    &  31.52 &  31.55               &    31.52   &  \textbf{31.56}\\\hline
N2  &  29.52  &  27.71  &  28.27    &  31.16 &  \textbf{31.20}  &  31.17     &  \textbf{31.20}\\\hline
N3  &  22.97  &  21.95  &  22.12    &  23.94 &  \textbf{24.00}  &   23.80 &  23.94\\\hline
N4  &  21.63  &  20.98  &  20.90    &  22.59 &  \textbf{22.66}  &    22.38  &   22.41\\\hline
N5  &  24.85  &  23.85  &  24.01    &  26.01 &  \textbf{26.06}  &  25.77   &  25.90\\\hline
N6  &  25.34  &  24.61  &  24.70    &  26.20 &  \textbf{26.26}  &    26.08  &   26.07\\\hline
N7  &  26.66  &  25.43  &  25.73    & 27.92  &    27.92             &  27.77   &  \textbf{27.97}\\\hline
N8  &  26.08  &  24.71  &  25.23    &  27.27 &  \textbf{27.43}  &   27.01  &    27.26\\\hline
N9  &  26.02  &  25.29  &  25.42    &  26.82 &  \textbf{26.89}  &  26.58  &    26.73\\\hline
N10  &  24.79  &  24.07  &  23.92  &  26.23 &  \textbf{26.25}  &  25.91  &  26.16\\\hline
N11  &  26.86  &  25.22  &  25.97  &  28.06 &  28.04  &   27.99  &    \textbf{28.16}\\\hline
N12  &  28.16  &  26.65  &  27.07  &  29.63 &  29.66  &   29.78  &    \textbf{29.86}\\\hline
N13  &  25.15  &  24.18  &  24.22  &  26.40 &  \textbf{26.36}  &    26.31  & 26.33\\\hline
N14  &  26.82  &  25.98  &  25.92  &  27.99 &  \textbf{28.01}  &    27.86  &  27.94\\\hline
N15  &  25.78  &  24.64  &  24.81  &  27.00 &  27.04  &   26.90  &   \textbf{27.06}\\\hline
N16  &  27.28  &  25.85  &  26.16   &  28.88 &  \textbf{29.01}  &  28.83  &   28.96\\\hline
N17  &  27.79  &  26.33  &  26.81   &  29.21 &  \textbf{29.24}  &    29.02  & 29.16\\\hline
N18  &  29.13  &  27.75  &  28.18  &  30.38 &  30.41  &   30.25  &  \textbf{30.43}\\\hline
N19  &  24.57  &  23.19  &  23.50  &  26.07 &  \textbf{26.10}  &    25.92  &   26.02\\\hline
N20  &  22.00  &  21.13  &  21.05  &  23.26 &  23.28  &    23.26  &  \textbf{23.29}\\\hline
  \end{tabular}     
   \label{tab:benchresults2}
    \end{center}
 \end{table}
 
As a quantitative measure of performance we compute the signal to noise ratio (PSNR), a measure that is widely used in image recovery applications. We present the PSNR results for benchmark images in Table \ref{tab:benchresults} and natural images in Table \ref{tab:benchresults2}. These PSNR based results can be summarized as: (1) The online learning algorithm and ScSR performs similarly, (2) They both slightly perform better than the Gibbs sampler. (3) All of the example based algorithms perform better than the interpolation based techniques.

\subsection{Evaluation  and Crowdsourcing via Mechanical Turk}
Though signal to noise ratio (PSNR), is a widely used metric in image recovery applications, this is not enough to measure human judgement. For this purpose, we also performed human evaluation experiments on Amazon Mechanical Turk (MTurk, http://www.mturk.com).

\begin{figure}
\begin{center}
\subfigure{\includegraphics[scale = 0.5]{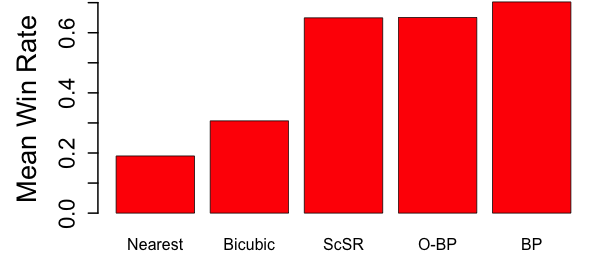}}
\subfigure{\includegraphics[scale = 0.5]{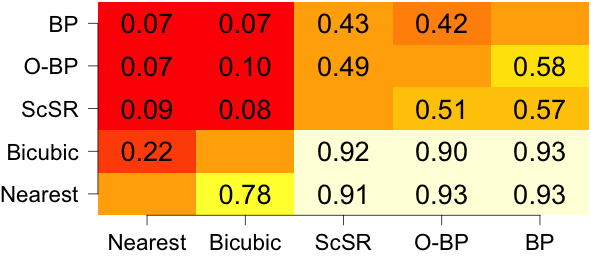}}
\end{center}
\caption{  \textbf{Human Evaluation via Mechanical Turk}.   (Top) Average win rate in one-to-one comparisons. (Bottom) Win rates for each one-to-one comparison. Each  number represents the winning rate of the method in the column, e.g., $0.57$ for BP vs ScSR (BP is on the column and ScSR on the row) means that on average, 0.57 of the times Turkers voted in favor of BP. }
\label{Mturk}
\end{figure}

The Amazon Mechanical Turk (MTurk) is a web interface for deploying small tasks to people, called \textit{Turkers}.  Typically an MTurk experiment works as follows: the \textit{requesters}, people organizing the experiments and paying Turkers, prepare tasks called HITs (Human Intelligence Tasks).  Each HIT might be a comparison of images, labeling of text etc. Once the HITs are completed, requesters can approve or reject the HITs based on their reliability measures (for instance trivial solution HITs, as we explain next, and the time spend on each HIT are frequently used measures for reliability). Approved results are acquired to be used in the analysis.

\begin{figure}
\begin{center}
\subfigure[High]{\includegraphics[scale = 1]{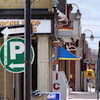}}
\subfigure[Low]{\includegraphics[scale = 1]{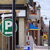}}
\subfigure[Bicubic]{\includegraphics[scale = 1]{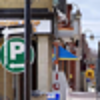}}
\subfigure[NNI]{\includegraphics[scale = 1]{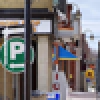}}
\subfigure[Bilinear]{\includegraphics[scale = 1]{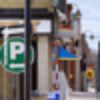}}
\subfigure[ScSR]{\includegraphics[scale = 1]{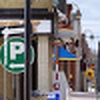}}
\subfigure[BP]{\includegraphics[scale = 1]{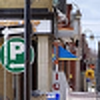}}
\subfigure[O-BP]{\includegraphics[scale = 1]{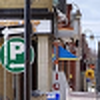}}
\end{center}
\caption{ \textbf{Reconstruction of Natural Image 3}. \textbf{BP}: Algorithm presented in this work  trained via Gibbs sampler, \textbf{O-BP}  Algorithm presented in this work  trained via Online VB,   \textbf{ScSR}: Super-Resolution via Sparse Representation. Example based approaches are superior to interpolation techniques, ScSR and our approach perform similarly. }
\label{Test3.jpg}
\end{figure}

While preparing HITs, we used the natural image data.  We asked Turkers to visually assess and select the better of two HR reconstructions of each image. We considered all ordered combinations of the algorithms, each equally likely, e.g.,  BP vs ScSR, BP vs Bicubic etc.  We initially collected $42,807$ decisions from  208 unique Turkers. For quality control we gave test pairs in which a ground truth HR image was used, i.e., a comparison of an algorithmic reconstruction vs a true HR image.  All of the judgments of the Turkers who failed to pass this test (Turkers who selected the algorithmic reconstruction instead of true HR) were removed.  This reduced the data to $20,469$ decisions from 161 unique reliable Turkers. 

The results of
the human evaluation are in Figure~\ref{Mturk}. In the bottom table, win rates for each one-to-one comparisons are provided. Each  number represents the winning rate of the method in the column. For instance, $0.93$ for O-BP vs Nearest (O-BP is on the column and Nearest on the row) means that out of 100 binary comparisons of O-BP and Nearest, 93 of the times Turkers voted in favor of O-BP.  In general, we observe that example-based methods perform significantly better than interpolation-based methods. Within the example-based approaches, the models are similar. However, our approach does not use the first and second-order derivative filters for the LR patches used by ScSR as features, yet we perform similarly; moreover we do not need to set the noise precision and the number of dictionary elements, both required parameters of ScSR (We used the parameters provided by \cite{superresolution1} in ScSR.).

In the PSNR results, ScSR and O-BP seem to perform similarly and both slightly better than BP. However, in the human evaluation we observed that BP reconstructions are found to be better.  (Based on 95\% confidence intervals, both the BP vs O-BP and BP vs ScSR results are statistically significant. The O-BP vs ScSR difference is statistically insignificant.)  This shows that PSNR is not necessarily consistent with the human assessment of images \cite{wang2009mean}.  Sample visual results  are  shown in Figures  \ref{Test3.jpg}, \ref{Parthenon} and \ref{Baboon}. (The remaining results are in the appendix E and F.)

\begin{figure}
\begin{center}
\subfigure[High]{\includegraphics[scale = 1]{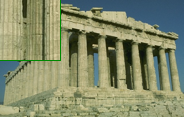}}
\subfigure[Low]{\includegraphics[scale = 1]{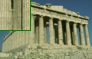}}
\subfigure[Bicubic]{\includegraphics[scale = 1]{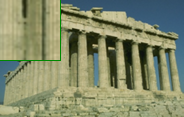}}
\subfigure[NNI]{\includegraphics[scale = 1]{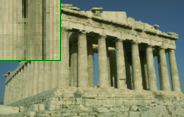}}
\subfigure[Bilinear]{\includegraphics[scale = 1]{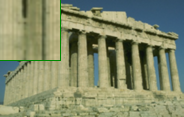}}
\subfigure[SME]{\includegraphics[scale = 1]{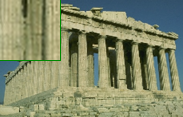}}
\subfigure[ScSR]{\includegraphics[scale = 1]{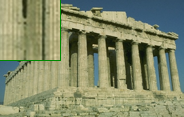}}
\subfigure[BP]{\includegraphics[scale = 1]{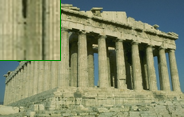}}
\subfigure[O-BP]{\includegraphics[scale = 1]{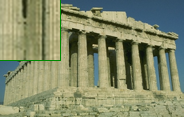}}
\caption{\textbf{Reconstruction of Parthenon Image}. \textbf{BP}: Algorithm presented in this work  trained via Gibbs sampler, \textbf{O-BP}  Algorithm presented in this work  trained via Online VB,   \textbf{ScSR}: Super-Resolution via Sparse Representation. \textbf{SME}: Sparse Mixing Estimation \cite{SME}}
\label{Parthenon}
\end{center}
\end{figure}

\begin{figure}
\begin{center}
\subfigure[High]{\includegraphics[scale = 0.92]{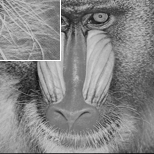}}
\subfigure[Low]{\includegraphics[scale = 0.92]{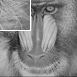}}
\subfigure[Bicubic]{\includegraphics[scale = 0.92]{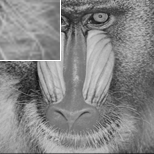}}
\subfigure[NNI]{\includegraphics[scale = 0.92]{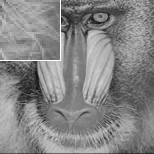}}
\subfigure[Bilinear]{\includegraphics[scale = 0.92]{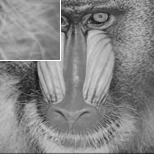}}
\subfigure[SME]{\includegraphics[scale = 0.92]{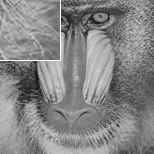}}
\subfigure[ScSR]{\includegraphics[scale = 0.92]{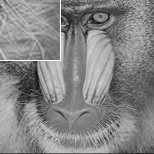}}
\subfigure[BP]{\includegraphics[scale = 0.92]{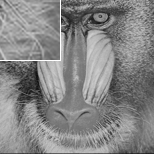}}
\subfigure[O-BP]{\includegraphics[scale = 0.92]{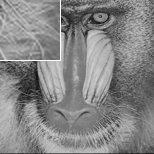}}
\end{center}
\caption{\textbf{Reconstruction of Baboon Image}. \textbf{BP}: Algorithm presented in this work  trained via Gibbs sampler, \textbf{O-BP}  Algorithm presented in this work  trained via Online VB,   \textbf{ScSR}: Super-Resolution via Sparse Representation. \textbf{SME}: Sparse Mixing Estimation \cite{SME}.}
\label{Baboon}
\end{figure}
\newpage

\subsection{Nonparametric property of the model.} 
In this section, we demonstrate the importance of a Bayesian nonparametric method for image super-resolution. As we mentioned in Section \ref{sec:BP}, we use a beta-Bernoulli process (BP) for the factor assignments $\textbf{z}_i$ that encodes which dictionary elements are activated for the corresponding observation.  In the binary matrix (whose rows are  the factor assignment $\textbf{z}_i$'s), the columns with at least one active cell correspond to factors that are used. 

\begin{figure}
\begin{center}
\subfigure{\includegraphics[scale = 0.44]{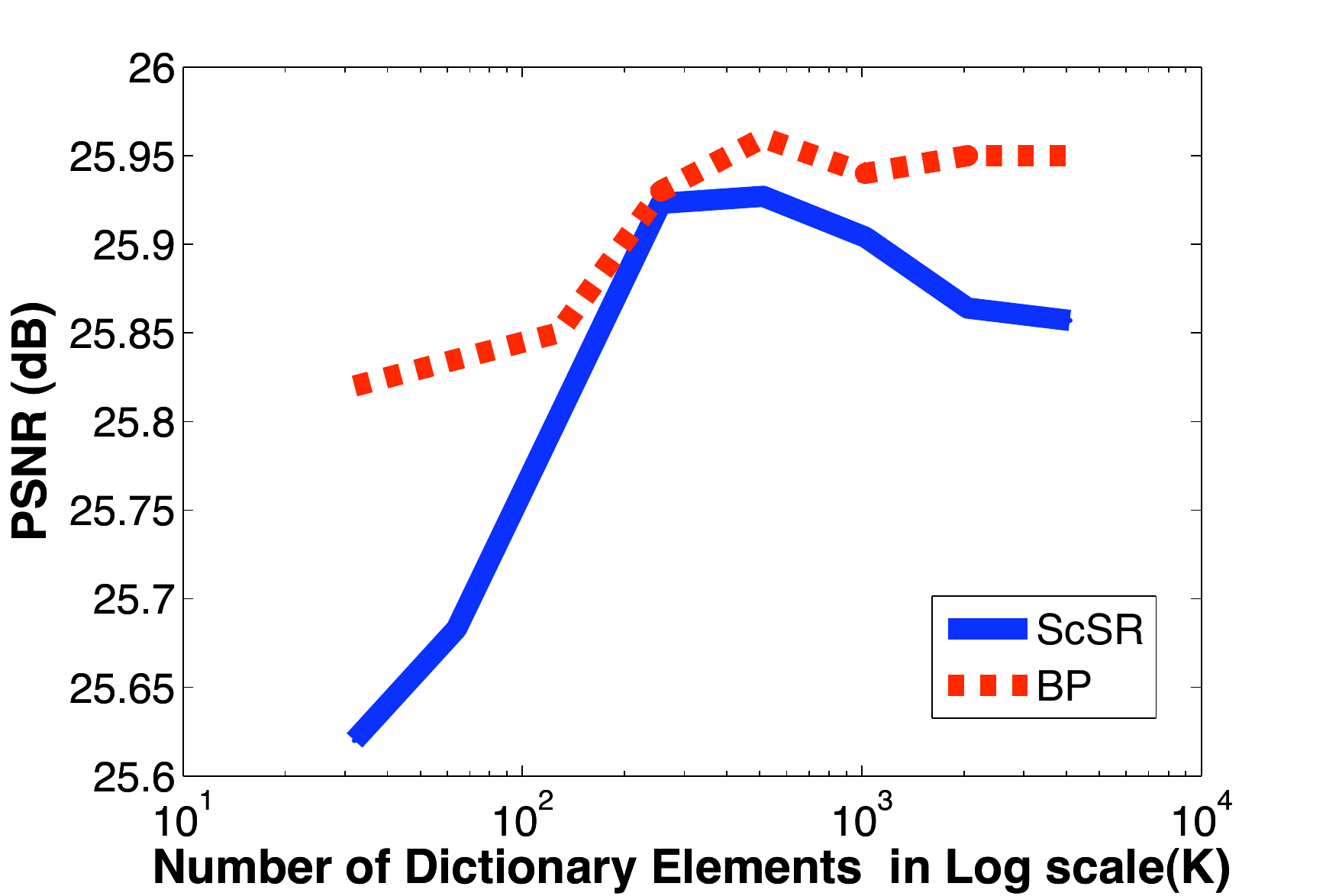}}
\subfigure{\includegraphics[scale = 0.44]{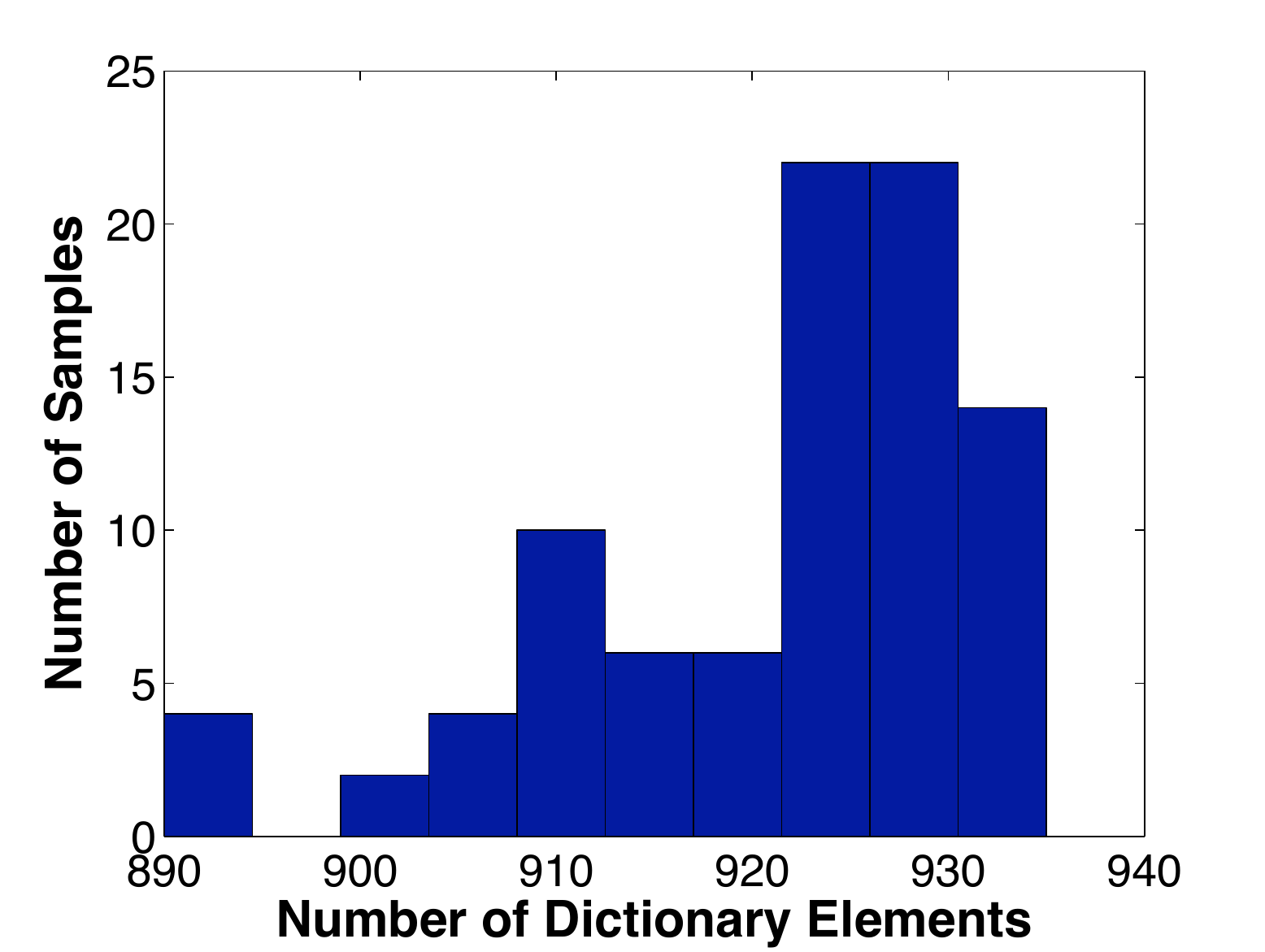}}
\end{center}
\caption{ \textbf{Learning the number of dictionary elements from the data}.   (Left) PSNR of the reconstruction of the Barbara image by nonparametric BP and parametric ScSR with different number of dictionary elements. (Right) Histogram of the number of dictionary elements for BP when $K=1024$ over 100 samples.  }
\label{nonparametric}
\end{figure}

The distinguishing characteristic of this prior is that the number of the factors to be learned is not specified a priori. Conditioned on the data, we examine the posterior distribution of the binary matrix to obtain a data-dependent distribution of how many components are needed. For the parametric  ScSR,   the number of dictionary elements must be set $\textit{a priori}$.  This is illustrated by the following experiment. For both model, we train on $10^4$ patches, for different values of $K$ (starting from scratch each time); for ScSR, $K$ is the target number (which needs to be set before starting the algorithm),  while for our approach, $K$ functions as an upper bound on the  number of dictionary elements (which should not be too low). Figure \ref{nonparametric} shows that, unlike ScSR, our approach is less sensitive to the value of $K$ if it is sufficiently large.  The Barbara image uses  700,  801 and  816 factors in our approach for  $K$ equals to $1024, 2048$ and $4096$ respectively.  %

\begin{figure}
\includegraphics[scale = 0.65]{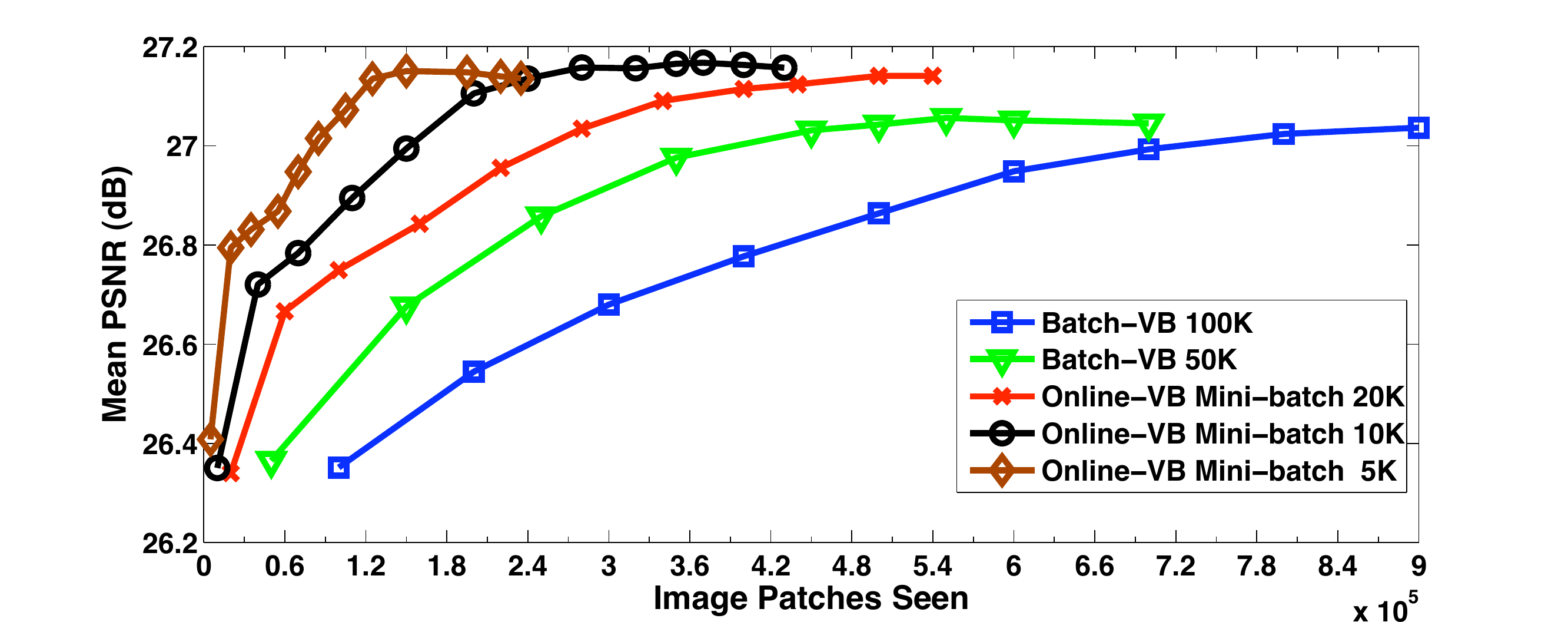}
\caption{\textbf{Held-out prediction performances of Online Learning} with different mini-batch sizes. Online-VB run on the whole data set is compared with  the Batch-VB run on a subset of the data.  The online algorithms converge much faster than the batch algorithm does.  }
\label{vbExperimentResults}
\end{figure}

\subsection{Online learning, Computational Time and Scaling}
\label{section:time}
In this section, we compare the scale properties of the algorithms presented in this paper. 
In online learning, instead of subsampling the patches during the dictionary learning
stage, we use the full data set and process the data segment by
segment (so called "mini-batches"). We use the training data set of
Section \ref{base_experiments}.  The learning parameters are set to
$\kappa = 0.501$ and $\rho_0 = 3$.  

 Figure \ref{vbExperimentResults}
shows the evolution of the mean PSNR on the held-out natural image
data set  by the online and the batch algorithms as a function of the
number of image patches seen (visualizations of the learned
dictionaries are provided in appendix D).  The number of patches seen represents the computational time since both algorithms' time complexity is linear with number of observations. For online VB, the number of patches seen represents the total number of data seen after each iteration. For batch VB, this represents cumulative sum of the number of same data seen after each variational-EM iteration. Even before the second iteration of the batch VB (100K) is completed, online VB with 5K mini-batch converges -- reaches to a local optima better than batch VB. This means that the online algorithm
finds dictionaries at least as good as those found by the batch VB in
only a fraction of the time. As also shown in Table
\ref{tab:benchresults}, it finds high quality dictionaries.  This may
be because stochastic gradient is robust to local optima \cite{Bottou98onlinelearning}.

For dictionary training,  the convergence time for online VB with 5K mini-batch size is 16 hours. In Gibbs sampling, we throw away the first 1500 samples for the burn-in period and later collect 1500 samples to approximate the posterior distributions. This takes approximately 50 hours on the same machine with an unoptimized Matlab implementation on $10^5$ number of patches. Running Gibbs sampling same amount of time with online VB, i.e. collecting less number of samples such as 500, reduces held-out PSNR between $0.2$ dB to $0.5$ dB, depending on the image. This is consistent with the findings in \cite{zhou2009non}.

\section{Discussion}
\label{discussion} %
We developed a new model for super-resolution based on Bayesian
nonparametric factor analysis, and new algorithms based on Gibbs
sampling and online variational inference.  With online training, our
algorithm scales to very large data sets.  We evaluated our method
against a leading sparse coding technique \cite{superresolution1} and other state of the art methods.  We evaluated both with traditional PSNR and by devising a large scale human evaluation. %
This is a new real-world application that can utilize online variational methods.

The choice of the inference algorithm  depends on the usage case. Our results suggest that with more computation time Gibbs sampling performs slightly better (based on human evaluation). If speed is important, our online algorithms can be used without much loss.

Regarding the evaluation metric, the standard in image analysis has been signal-to-noise ratio (PSNR).  However, its practical relevance has been questioned \cite{wang2009mean}.  The human eye is sensitive to details which are not always captured in this metric, and that is why we ran a human evaluation. Our experiments show that the signal-to-noise ratio is not necessarily consistent with human judgement. %

As future work, our approach can be used as a building block in other, more complicated, probabilistic models.  For example, our approach could be developed into a time series to perform super-resolution on video or  a  hierarchical model can be built  that fully generates the whole image instead of patch based approach.

\begin{table}
 \caption{Test Set Results with SR ratio = 4. PSNR for the illuminance channel (higher the better). \textbf{BP}: Algorithm presented in this paper with Beta Process (BP) Prior trained with Gibbs Sampler,   \textbf{ScSR}: Super-Resolution via Sparse Representation,  \textbf{NNI}: Nearest neighbor interpolation. }
   \begin{center}
  \begin{tabular}{| c | c | c | c | c |c |c |c |c |c |c |}
    \hline
  \textbf{PSNR} 			& Bicubic            & NNI      & Bilinear      &ScSR     & BP  \\\hline
N1  &  24.58  &  22.80  &  23.77  &  25.36  &  25.15\\\hline
N2  &  24.81  &  23.54  &  24.18  &  25.51  &  25.28\\\hline
N3  &  18.97  &  18.43  &  18.60  &  19.39  &  19.38\\\hline
N4  &  18.11  &  17.78  &  17.83  &  18.50  &  18.37\\\hline
N5  &  21.17  &  20.60  &  20.78  &  21.72  &  21.54\\\hline
N6  &  22.12  &  21.68  &  21.84  &  22.43  &  22.41\\\hline
N7  &  22.62  &  21.62  &  22.20  &  23.13  &  23.12\\\hline
N8  &  22.00  &  20.82  &  21.53  &  22.59  &  22.47\\\hline
N9  &  22.90  &  22.27  &  22.59  &  23.10  &  23.16\\\hline
N10  &  21.39  &  21.04  &  21.22  &  21.53  &  21.55\\\hline
N11  &  22.82  &  21.28  &  22.22  &  23.51  &  23.41\\\hline
N12  &  24.09  &  23.10  &  23.53  &  24.74  &  24.66\\\hline
N13  &  21.13  &  20.42  &  20.77  &  21.42  &  21.45\\\hline
N14  &  23.06  &  22.50  &  22.72  &  23.31  &  23.33\\\hline
N15  &  21.79  &  21.15  &  21.45  &  22.11  &  22.16\\\hline
N16  &  22.52  &  21.58  &  22.05  &  23.00  &  22.92\\\hline
N17  &  23.70  &  22.66  &  23.19  &  24.25  &  24.10\\\hline
N18  &  25.21  &  24.38  &  24.74  &  25.48  &  25.62\\\hline
N19  &  20.33  &  19.55  &  19.89  &  20.79  &  20.76\\\hline
N20  &  18.31  &  17.92  &  18.00  &  18.54  &  18.67\\
\hline 
  \end{tabular}
    \end{center}
    \label{tab:natresults2}
\end{table}

  \begin{figure}
\begin{center}
\subfigure[High]{\includegraphics[scale = 1]{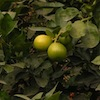}}
\subfigure[Low]{\includegraphics[scale = 1]{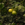}}
\subfigure[Bicubic]{\includegraphics[scale = 1]{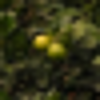}}\\
\subfigure[NNI]{\includegraphics[scale = 1]{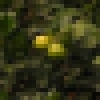}}
\subfigure[ScSR]{\includegraphics[scale = 1]{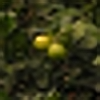}}
\subfigure[BP]{\includegraphics[scale = 1]{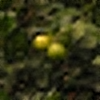}}
\end{center}
\label{Test18.jpg}
\caption{(Natural Image 18)  Test Set Results with SR ratio = 4.  }
\end{figure}

\begin{figure}
\begin{center}
\subfigure[High]{\includegraphics[scale = 1]{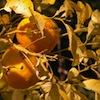}}
\subfigure[Low]{\includegraphics[scale = 1]{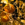}}
\subfigure[Bicubic]{\includegraphics[scale = 1]{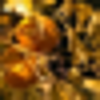}}\\
\subfigure[NNI]{\includegraphics[scale = 1]{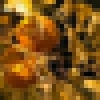}}
\subfigure[ScSR]{\includegraphics[scale = 1]{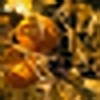}}
\subfigure[BP]{\includegraphics[scale = 1]{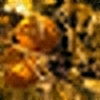}}
\end{center}
\label{Test19.jpg}
\caption{(Natural Image 19)  Test Set Results with SR ratio = 4.  }
\end{figure}

\bibliography{refs2}
\bibliographystyle{IEEEtran}
\end{document}